\documentclass[11pt]{article}

\usepackage[utf8]{inputenc}
\usepackage{lmodern}
\usepackage[T1]{fontenc}
\usepackage{amsmath,amssymb}
\usepackage[margin=1in]{geometry}
\usepackage[round,authoryear]{natbib}
\usepackage{hyperref}
\hypersetup{colorlinks=true, linkcolor=blue, citecolor=blue, urlcolor=blue}
\newcommand{\Lown}{L_i^{\mathrm{own}}}
\newcommand{\Lsoc}{L_i^{\mathrm{social}}}
\newcommand{\Veff}{V_i^{\mathrm{eff}}}
\newcommand{\Rred}{R_{\mathrm{red}}}
\newcommand{\E}{\mathbb{E}}
\newcommand{\Sset}{\mathcal{K}}

\title{A framework for single and multi-agent human-AI curiosity ecosystems}
\author{Ilya E. Monosov\\[4pt]
  {\small Solomon H. Snyder Department of Neuroscience, Johns Hopkins University, Baltimore, MD, USA}\\
  {\small Departments of Biomedical Engineering, Electrical and Computer Engineering, and Psychiatry,}\\
  {\small Johns Hopkins University, Baltimore, MD, USA}\\
  {\small Zanvyl Krieger Mind/Brain Institute, Johns Hopkins University, Baltimore, MD, USA}\\
  {\small Data Science and Artificial Intelligence Institute and the Kavli Neuroscience Discovery Institute,}\\
  {\small Johns Hopkins University, Baltimore, MD, USA}\\[4pt]
  \texttt{ilya.monosov@gmail.com}}
\date{}

\begin{document}
\maketitle

\section*{Highlights}

\begin{itemize}
  \item Asking a question is a choice among competing values in immediate uncertainty reduction, effort, delayed return, and the value of leaving the question open.
  \item The weights on those values can change or drift. Repeated exposure to fast cheap answers may gradually encourage fast resolution to be more attractive and long-horizon inquiry less attractive.
  \item The framework can help to generate future theories that can be applied to groups of people or AI agents that share and explore a knowledge landscape. Over time, it can also inform multi-agent AI development. 
  \item This work supports future studies of how inquiry becomes adaptive and generative, or maladaptive.
\end{itemize}

\begin{abstract}
This paper offers a framework for considering curiosity as an ecosystem. \textit{First}, it suggests that a single agent's inquiry policy (how, when, and why an agent asks a question) depends on how the agent values immediate uncertainty reduction, costs, delayed return, and the value of keeping the question open. A key concept in the framework is that the weights on these decision-related terms can change with experience. For example, a period of cheap, quickly answered questions may change the cost of inquiry on a short timescale and change which kinds of questions the agent is drawn to answer over a longer timescale. \textit{Second}, these ideas are extended to many agents exploring a shared knowledge landscape, and there the framework tracks inquiry volume, topic diversity, frontier-directed inquiry, redundancy, and reusable knowledge. The result is a conceptual framework for studying curiosity ecology and for future efforts towards designing multi-agent AI systems for discovery.
\end{abstract}

\section{Introduction}

Why do we ask certain questions and avoid asking others? A question can tempt us because the answer will remove lots of uncertainty, because the answer may pay off substantially, even if only later, or because it is easy to pursue, or simply because everyone around us is already asking it. Models of curiosity and information-seeking suggest that the subjective value we place on an answer rises and falls with how uncertain we are and how much uncertainty we think the answer will resolve \citep[for review see][]{monosov2024,brombergmartin2020circuitry}. This subjective value of information can change on a moment-by-moment basis. 

A lasting change in context can change how our preferences are expressed, or change the preferences themselves \citep{tversky1993}. One neurobiologically inspired way to formalize this is to assume that the brain has weights for different features or attributes of a question (e.g., amount of uncertainty it would reduce, time to ask, cost, delayed payoff, and so on) and to allow the weights to move, drift, or be updated systematically in response to context and experience. Then, the questions a person has already been asking (their history), the cost of the tools available, the questions being asked by their peers, and the value attached to answers in the surrounding environment shape what questions feel like they are worth pursuing next. The same question can look trivial in one ecology or one agent and valuable in another ecology (to the same or to other agents).

This paper attempts to consider this complexity by linking three processes that are often treated separately: the process through which a single agent chooses to pursue or not to pursue a question, the drift in that agent's inquiry preferences that occurs as a function of their own experience and context, and the process through which many agents build up a shared stock of knowledge and express curiosity collectively. The resulting toy framework can be helpful in considering the adaptivity of a curiosity ecosystem and for future work designing adaptive multi-agent AI systems.

\section{Single agent's inquiry}\label{sec:single}

Consider a single agent $i$ deciding whether to pursue question $q$ at time $t$. The agent's private value for that question is
\begin{equation}\label{eq:value}
V_i(q,t) = \psi_i(t)\, \E_{i,t}[I_i(q,t)] - \lambda_i(t)\, \E_{i,t}[C_i(q,t)] + \tau_i(t)\, \E_{i,t}[L_i(q,t)] - \phi_i(t)\, \E_{i,t}[O_i(q,t)].
\end{equation}
Here $I_i$ is the immediate value of reducing uncertainty. $C_i$ is the cost of asking (e.g., effort, time, attention). $L_i$ is the long-horizon return, meaning what the answer may become useful for later. $O_i$ is the value of not answering yet. That is the value of keeping the question open, either for flexibility \citep{mcdonald1986,dixit1994} or because some information is actively not desired by the agent \citep{golman2017,gigerenzer2017}. This is further inspired by neural data that suggest that information about rewards and punishments can be valued differently, with partly distinct circuits for information seeking and information avoidance \citep{jezzini2021,charpentier2018}. $O_i$ is meant to capture cases in which an agent would decline to resolve $q$ right now even when it could be tempting. A scientist may keep a promising hypothesis deliberately open to preserve the flexibility of running a more informative experiment, or a person may prefer not to learn a medical or genetic result. In these examples, keeping the question open is distinct from, and can even act against, the long-horizon return $L_i$. 

The weights $\theta_i(t) = (\psi_i(t), \lambda_i(t), \tau_i(t), \phi_i(t))$ reflect how much the agent cares about each term and define its \emph{curiosity policy}. All four are written as expectations at time $t$ ($\E_{i,t}[\cdot]$), prior to asking $q$. Looking up who won a chess tournament last night is mostly driven by $I_i$: it is expected to resolve uncertainty immediately and costs almost nothing. Working through a hard homework problem is costly now, so $C_i$ is high, but the delayed return $L_i$ may be high as well.

$I_i$ and $L_i$ are defined by how many questions must be resolved for their value to be realized. $I_i$ is the myopic value of resolving $q$ on its own as if no further question were asked; this value can arrive far in the future and still count as $I_i$ so long as it depends only on $q$. $L_i$ is instead the value that requires the follow-up questions opened by $q$ to be resolved. The long-horizon term can be split into private and social returns:
\begin{equation*}
L_i(q,t) = \Lown(q,t) + \kappa_i \Lsoc(q,t).
\end{equation*}
The first component is the future payoff that is collected by the agent itself. The second is the payoff that spills over to other agents, discounted by $\kappa_i \in [0,1]$. A self-interested agent has $\kappa_i=0$; an agent that cares about downstream benefits to others has $\kappa_i>0$. 

A second important property of a question is its generativity $g_i(q,t)$: the number of new knowledge gaps that resolving $q$ opens, relative to what the agent already knows. This is a gross count of the gaps that $q$ opens (the one gap that $q$ closes by being answered is not deducted from it, so a question that opens even a single new gap has $g_i>0$). The simple intuition for this definition is that learning can make a person aware of what they don't know  \citep{loewenstein1994,murayama2022}. For example, a trivia question usually closes its own knowledge gap and often opens little opportunity for future uncertainty reduction, so $g_i \approx 0$. A question embedded in a sparse region of the knowledge landscape can do the opposite. For example, new scientific data can answer an immediate question and open an entirely new research program. 

For the agent's own return, a simple specification is
\begin{equation*}
\Lown(q,t) = \ell_0 + \rho_i \cdot g_i(q,t),
\end{equation*}
where $\ell_0$ is a baseline return that is constant across questions, and $\rho_i$ converts newly opened knowledge gaps into expected future payoff. 

Note that this is not the same as $O_i$ in Equation~\ref{eq:value}. $O_i$ is the value of not closing a question right now. In contrast, $g_i$ counts the new questions that appear once a question\textit{ is} closed. 

Also note that a cheap question high in $I_i$ can be near zero in $g_i$. Such a question could feel good to answer now (due to large uncertainty reduction), but it would not create future opportunities for more uncertainty reduction or learning. A frontier question (that is one that is difficult but may produce discoveries) may be less satisfying in the moment and more costly to pursue, but can be high in $g_i$ and therefore high in $L_i$. Which one is chosen depends on the relative weight placed on immediate resolution and long-horizon return, especially $\psi_i$ and $\tau_i$.

The social component $\Lsoc$ can be specified the same way, with the group in place of the individual. It depends on the question's \emph{population generativity} $g_s(q,t)$: the number of new knowledge gaps it opens relative to what the group already knows rather than to one agent's knowledge (treated in Section~\ref{sec:population}). There are two ways to specify it. At the micro level, the social return is the spillover that $i$'s resolution of $q$ creates for everyone else,
\begin{equation*}
\Lsoc(q,t) = \sum_{j\neq i} \Delta L^{\mathrm{own}}_{j}(q,t),
\end{equation*}
where $\Delta L^{\mathrm{own}}_{j}(q,t)$ is the long-horizon own-return that $i$'s answer confers on agent $j$. But this requires knowing every pairwise spillover. At the macro level, when those pairwise terms are not observable, the same concept is summarized through population generativity,
\begin{equation*}
\Lsoc(q,t) = \ell_0^{\mathrm{soc}} + \rho_i^{\mathrm{soc}}\, g_s(q,t),
\end{equation*}
where $\rho_i^{\mathrm{soc}}$ converts group-level gaps opened into expected downstream payoff to others (the social counterpart of $\rho_i$, and likewise agent-specific because agents differ in how far their answers propagate). 

The agent is more likely to pursue higher-value questions. In the framework, this can be implemented with a standard pursuit probability. Let $P_i(q,t)$ be the probability that agent $i$ pursues $q$ at time $t$, rising with $V_i(q,t)$ under a threshold rule, a logistic rule, or a softmax with an outside option (a fixed no-pursuit alternative included in the choice set, so that the agent may also choose to ask nothing). $P_i(q,t)$ is then a per-question pursuit probability. An agent can pursue several questions, or none, and the total volume of inquiry $Q_t$ is free to rise or fall.

\section{Preference drift}\label{sec:drift}

A new tool or a change in social settings can change Equation~\ref{eq:value} by changing the terms themselves, especially by lowering the cost $C_i$. However, over longer timescales, repeated inquiry may also change the weights themselves. A person repeatedly rewarded by fast answers may gradually become biased toward that kind of question. 

Recent inquiry is summarized by an experience state $X_i(t)$, a smoothed version of the profile $x_i(t)$ of questions the agent has resolved:
\begin{equation*}
x_i(t) = [\,\text{cost},\ \text{speed},\ \text{uncertainty reduction},\ \text{generativity},\ \text{reusability}\,]^{\top}.
\end{equation*}
Persistence $\gamma_X \in [0,1]$ controls how slowly past experience fades, inspired by economic models in which current preferences depend on past consumption \citep{ryder1973,becker1988}:
\begin{equation}\label{eq:exp}
X_i(t) = \gamma_X\, X_i(t-1) + (1 - \gamma_X)\, x_i(t).
\end{equation}
If no question is resolved, set $X_i(t)=X_i(t-1)$. The agent updates only from the questions it actually pursued and resolved. Question selection shapes experience, and experience then feeds back onto selection. And, the weights update through a drift operator:
\begin{equation}\label{eq:drift}
\theta_i(t+1) = \theta_i(t) + \Delta\theta_i\big(X_i(t), M_t\big).
\end{equation}
Here $M_t$ is the surrounding inquiry ecology. It includes the cost and tool landscape, the questions and values circulating in the population, and the shared knowledge stock $S_t$ already available (Section~\ref{sec:multiagent}). The drift term $\Delta\theta_i$ describes how a given region of this ecology nudges the weights $(\psi_i,\lambda_i,\tau_i,\phi_i)$ while keeping them in their natural nonnegative range. A curiosity regime is a region of the joint space of $X_i(t)$ and $M_t$. For example, in a cheap-inquiry regime, fast and low-cost resolution is abundant. The same history $X_i(t)$ can have different effects under different ecologies. One plausible implementation of the drift is reinforcement learning. Quickly answered questions can act as small, frequent subjective rewards. Over longer timescales, such rewards may reshape the value function that guides inquiry, for example through direct changes in valuation circuits \citep{sutton2018,schultz1997}. 

Different forms of $\Delta\theta_i$ could be useful for future simulations. One example is a linear update with decay,
\begin{equation}\label{eq:driftlinear}
\Delta\theta_i\big(X_i(t), M_t\big) = \varepsilon\,\big(W\,X_i(t) + w_M\, m(M_t) - \theta_i(t)\big),
\end{equation}
where $\varepsilon>0$ is a learning rate, $W$ maps the experience profile $X_i(t)$ onto target weights, $m(M_t)$ is a vector summary of the ecology, and $w_M$ scales its influence. The weights are clipped at zero after each update to keep $(\psi_i,\lambda_i,\tau_i,\phi_i)$ in their natural nonnegative range. Here $\theta_i$ relaxes toward a target set jointly by recent experience and the surrounding ecology, and the regime-dependent sign patterns (discussed below) correspond to the signs of the relevant entries of $W$ and $w_M$, and can flip across ecologies through the ecology summary $m(M_t)$. 

Other update rules, such as reinforcement learning, can be substituted without changing the rest of the framework.

Whether repeated cheap inquiry makes the agent \emph{more} or \emph{less} effort-averse---the sign of $\Delta\lambda_i$---depends on the ecology $M_t$. In terms of Equation~\ref{eq:driftlinear}, it depends on which channel dominates the drift: recent experience, $W\,X_i(t)$, or the surrounding ecology, $w_M\, m(M_t)$.

Consider first a \emph{habituation regime}, in which experience has a particularly strong influence on the agent's policy. The agent keeps receiving cheap answers that are low in generativity and reusability. Quick, low-cost resolution gets rewarded, so effort tolerance erodes and the agent becomes \emph{more} cost-sensitive:
\begin{equation*}
\Delta\psi_i > 0, \quad \Delta\lambda_i > 0, \quad \Delta\tau_i < 0, \quad \Delta\phi_i < 0.
\end{equation*}
Each cheap answer makes the next one more attractive. Now consider an \emph{abundance regime}, in which the broad ecology---rather than the agent's own recent experience---dominates the agent's preference drift. The agent has only a limited budget of effort to spend, but sits in an ecology where almost every question is fast and cheap and the costly, high-value questions are rare. Because those cheap questions consume so little of the budget, effort is no longer the binding constraint on inquiry: the agent can afford to take on each expensive frontier question it occasionally encounters and collect its high return. If $\lambda_i$ tracks how scarce effort feels, this abundance of cheap answers lowers it, and the agent becomes \emph{less} cost-sensitive:
\begin{equation*}
\Delta\psi_i > 0, \quad \Delta\lambda_i < 0, \quad \Delta\tau_i < 0, \quad \Delta\phi_i < 0.
\end{equation*}
The long-horizon weight $\tau_i$ still falls, because the experience state remains dominated by quick rewards. But the cost channel now works the other way. A lower $\lambda_i$ makes the costly frontier question easier to accept, even as the rising $\psi_i$ pulls toward the cheap one. The two channels are in tension, and the signs alone no longer settle the outcome.

To see which way the balance tips, let the agent choose between a cheap question $c$ and a frontier question $f$. Write $\Delta I = I_c - I_f$, $\Delta C = C_c - C_f$, and $\Delta L = L_c - L_f$; I ignore the $O_i$ term for simplicity. The cheap question resolves more uncertainty now ($\Delta I>0$), costs less ($\Delta C<0$), and pays less later ($\Delta L<0$). The drift strengthens the pull toward the cheap question only when
\begin{equation}\label{eq:loop}
(\Delta I)\,\Delta\psi_i + (\Delta L)\,\Delta\tau_i \;>\; (\Delta C)\,\Delta\lambda_i.
\end{equation}
The left side is the gain in attraction to quick resolution, plus the weakened pull of future payoff. The right side is the change in the cost term. In the habituation regime, $\Delta\lambda_i>0$ makes the right side negative, so the inequality holds whenever the drifts take the signs above. In the \emph{abundance} regime, $\Delta\lambda_i<0$ makes the right side positive, so the cost term works \emph{against} the pull toward cheap: shallow inquiry wins only if the immediacy channel outweighs it. The same history of cheap answers can therefore push an agent toward shallow inquiry in one ecology but not in another.

The model does not predict that cheaper access to answers always degrades curiosity. For example, if tools free up our time to pursue harder questions, or if cheap answers themselves open new knowledge gaps, the weight on long-horizon return $\tau_i$ may not drop. The drift in Equation~\ref{eq:drift} should not automatically be read as a change in deep preference \citep{stigler1977}. It may also reflect an altered environment or altered expression of a policy. Also, the drift is regulated by the kind of questions asked, not just by the number of questions. Two people can ask the same number of questions in a day and drift in opposite directions. Finally, the current examples emphasize cost and speed, but the same update rule can be applied to generativity, that is, to whether an answer opened new questions and created reusable by-products rather than only resolving a single question. 

\section{Multiple agents and shared knowledge}\label{sec:multiagent}

Discoveries are not solitary. People and machines explore a shared knowledge landscape and borrow from and duplicate one another, and one agent's work can make another agent's next question cheaper. This is a version of the exploration--exploitation problem \citep{nelson1982,march1991,averbeck2015,costa2020}, except that here one agent's answer can change both the value and the cost of another agent's inquiry.

To include this, each agent values a question along two dimensions: its private value, as in Equation~\ref{eq:value}, and what it may contribute to a shared stock of knowledge. Formally, the shared stock has two layers: a set $\Sset_t$ of resolved, deposited results, against which redundancy $d(q,\Sset_t)$ and reusable value $U(q,\Sset_t)$ are evaluated, and a scalar index $S_t$ of total reusable value, which evolves according to Equation~\ref{eq:stock}. The effective value of a question is
\begin{equation}\label{eq:eff}
\Veff(q,t) = V_i(q,t) + \eta_i\, \E_{i,t}[U(q,\Sset_t) \mid q\ \text{resolved}] - \zeta_i \Rred(q,t).
\end{equation}
The middle term rewards a reusable contribution to $\Sset_t$, scaled by $\eta_i$. Here $U(q,\Sset_t)\ge 0$ is the reusable, non-redundant value that resolving $q$ adds to the shared knowledge stock: value that other agents can later build on or reuse rather than having to reproduce, such as a result, tool, or dataset that lowers the cost of, or opens, their subsequent questions (a result whose content is already in $\Sset_t$ contributes $U=0$). It is specified in Equation~\ref{eq:stock}. The last term penalizes redundancy, scaled by $\zeta_i$. These two terms act at a different stage than the social benefits inside $V_i$. The long-horizon social return $\Lsoc$ is a delayed spillover to others, discounted by $\kappa_i$. In contrast, $\eta_i$ is an immediate reward for depositing a reusable result, and $\zeta_i$ is a penalty for duplicating existing or concurrent work. 

A redundancy score is
\begin{equation*}
\Rred(q,t) = \omega_d\, d(q,\Sset_t) + \omega_p \sum_{j\neq i} P_j(q,t),
\end{equation*}
where $d(q,\Sset_t)$ measures overlap with what is already known, and the sum measures how much other agents are already pursuing the same question. The nonnegative weights $\omega_d,\omega_p\ge 0$ convert these two distinct sources of overlap onto a common value scale; Section~\ref{sec:extensions} discusses setting them separately, including $\omega_p<0$ when parallel replication is desirable. These terms can affect the choice of a question, the decision to preserve a result in $\Sset_t$, or both.

The scalar index of shared knowledge evolves as an expected index of reusable knowledge:
\begin{equation}\label{eq:stock}
S_{t+1} = (1-\delta)\,S_t + \sum_q \Big[\,1 - \prod_i \big(1 - P_i(q,t)\, r_i(q,t)\big)\Big] \; U(q,\Sset_t),
\end{equation}
where $\delta\in[0,1]$ is a staleness rate, $r_i(q,t)$ is the probability that pursuit succeeds, and $U(q,\Sset_t)\ge 0$ is the reusable non-redundant value of resolving $q$ (results whose content is already in $\Sset_t$ produce $U(q,\Sset_t)=0$). Each resolved question also deposits its result into the set $\Sset_t$, so the two layers advance together.

The bracketed term is the probability that at least one agent resolves $q$, under the assumption that agents' pursuit and success events are independent conditional on the current state; correlated strategies would require replacing it with the joint probability that at least one agent resolves $q$. Each contribution is therefore credited once, even if several agents pursue the same question. The same $U(q,\Sset_t)$ also enters the agent's own value in Equation~\ref{eq:eff}, but only for the questions that agent resolves. The shared stock grows from everyone's contributions, but the reward for depositing knowledge goes only to the agent who contributed it.

In this framework, agents can be coupled in several ways. First, the drift operator in Equation~\ref{eq:drift} can read a peer summary $\bar{X}_{-i}(t)$, so an agent's weights are shaped by its own history together with what others are asking. That coupling can coordinate a population, but it can also make its questions more alike. Second, one agent's tools, explanations, or partial answers can lower the cost $C_i$ for another. Third, the shared stock $S_t$ grows only from contributions that are reusable and non-redundant, and it decays as knowledge becomes stale. Thus the same amount of activity can leave behind very different bodies of knowledge.

Simply adding more agents may not be enough to grow the knowledge ``stock''. A group with the same goal, the same evidence, and the same starting point can generate many parallel inquiry trajectories that converge on effectively the same knowledge rather than genuinely diverse results. This is a concern for large language models trained on similar data and decoded in similar ways, where more samples do not necessarily mean more production of functional diversity \citep{doshi2024,padmakumar2024,shumailov2024}. To make discoveries, a multi-agent system needs three things: division of labor, a shared state that agents can inspect, and outputs passed to others in a reusable format. The parameters $\eta_i$ and $\zeta_i$ help with discovery but they cannot replace or outpace this structure. $\eta_i$ and $\zeta_i$ are only meaningful if the agent can evaluate the quantities they act on, $\E_{i,t}[U(q,\Sset_t)]$ and $\Rred(q,t)$. Computing those requires access to the shared knowledge: what is already known and what peers are pursuing. These parameters therefore presuppose that structure rather than substituting for it. They are most critical when agents are scarce relative to the knowledge landscape (e.g., there are more open questions than agents or when the landscape is too large or even dangerous for uncoordinated search).

\section{Population-level quantities}\label{sec:population}

At the population level, the analog of a single agent's generativity $g_i$ is a question's ability to open knowledge gaps relative to what the group already knows. In a simplified setup, this is $g_s(q,t)$, evaluated against the shared knowledge stock $S_t$; it is the driver of the social long-horizon return $\Lsoc$ specified in Section~\ref{sec:single}. Because a generative question often benefits (or should benefit) the group more than the individual who asks it \citep{arrow1962}, agents with low $\kappa_i$ tend to under-pursue such questions.

Generative questions tend to add reusable value to the shared stock through $\E_{i,t}[U(q,\Sset_t)\mid q\ \text{resolved}]$. Questions whose generativity $g_s$ exceeds a threshold $\bar{g}$ make up the frontier set $\mathcal{F}_t$:
\begin{equation*}
\mathcal{F}_t = \{q : g_s(q,t) \geq \bar{g}\}.
\end{equation*}

To study a population's contribution to knowledge, I use three quantities. First, query volume is the total inquiry pursuit mass,
\begin{equation*}
Q_t = \sum_i \sum_q P_i(q,t).
\end{equation*}
Second, topic diversity measures how evenly that mass is spread across $K$ topics $\omega$, where each topic $\omega$ is a mutually exclusive set of questions and the $K$ topics together partition the space of questions.
\begin{equation*}
p_t(\omega) = Q_t^{-1} \sum_i \sum_{q\in\omega} P_i(q,t),
\end{equation*}
which I quantify using the normalized Shannon entropy \citep{shannon1948}
\begin{equation}\label{eq:div}
D_t = -\frac{1}{\log K} \sum_{\omega=1}^{K} p_t(\omega)\, \log p_t(\omega).
\end{equation}
This equals $0$ when all inquiry sits on one topic and $1$ when inquiry is evenly distributed, using the convention $0\log 0=0$ and assuming $K>1$. $D_t$ measures spread across topics but not semantic distance between them. 

The third quantity is the frontier share defined as the fraction of inquiry aimed at $\mathcal{F}_t$:
\begin{equation}\label{eq:fshare}
F_t = \frac{\sum_i \sum_{q \in \mathcal{F}_t} P_i(q,t)}{\sum_i \sum_q P_i(q,t)}.
\end{equation}
Returns to inquiry are often heavy-tailed, so a small number of questions can account for much of the eventual impact \citep{uzzi2013,foster2015}. If $Q_t=0$, I set $D_t=F_t=0$ and read the population-performance index $R_t^S$ (defined in Equation~\ref{eq:cobb} below) as $R_t^S=0$ by convention. This index summarizes performance at the level of the whole population and is distinct from the per-question redundancy score $\Rred$ of Section~\ref{sec:multiagent}.

The next step summarizes, in a single number, how effectively a population's inquiry produces valuable knowledge. The three quantities are combined into a population-performance index $R_t^S$ that is high only when inquiry is simultaneously substantial in volume, spread across topics, and aimed at the frontier, rather than maximizing one of these at the expense of the others. The index is inspired by a Cobb--Douglas form \citep{cobb1928}:
\begin{equation}\label{eq:cobb}
R_t^{S} = A^{S}\, Q_t^{a_Q}\, (\epsilon_D + D_t)^{a_D}\, (\epsilon_F + F_t)^{a_F},
\end{equation}
where $A^{S}>0$ is a scale constant (a total-factor-productivity term that sets the units and baseline efficiency of the index), and the exponents $a_Q,a_D,a_F\geq 0$ set the importance of volume, diversity, and frontier share. The small floors $\epsilon_D,\epsilon_F>0$ keep the index from collapsing to zero when a population sits on a single topic ($D_t=0$) or entirely off the frontier ($F_t=0$), while preserving the complementarity between the three terms. $D_t$ and $F_t$ are separated and can move in opposite directions. A population can spread inquiry across many topics without reaching the frontier, or concentrate on a few topics that are mostly frontier. More inquiry only raises performance when the added volume is not outweighed by losses in diversity or frontier-directed effort. This follows from Equation~\ref{eq:cobb}: for positive $Q_t$, $D_t$, and $F_t$, small proportional changes satisfy
\begin{equation*}
\Delta\log R_t^S = a_Q\,\Delta\log Q_t + a_D\,\Delta\log D_t + a_F\,\Delta\log F_t,
\end{equation*}
so a gain in volume ($a_Q\,\Delta\log Q_t$) helps net only if it is not cancelled by drops in the diversity or frontier terms.

\section{Extensions for multi-agent AI design}\label{sec:extensions}

The equations in this section and the one before it can serve as design ideas for multi-agent AI systems and can be adapted as needed. If a swarm needs to focus on a single objective, setting $a_D=0$ removes topic diversity from $R_t^S$. Annealing the pursuit rule $P_i(q,t)$ over time moves agents from exploration toward exploitation. Weighting the peer summary $\bar{X}_{-i}$ more heavily in the drift operator (Equation~\ref{eq:drift}) couples agents more tightly, coordinating the population but homogenizing the questions they ask.

Redundancy can sometimes be useful. The redundancy weight $\zeta_i$ can be made negative in cases in which replication is valuable rather than wasteful.  Re-deriving an answer already in the stock ($d(q,\Sset_t)$) wastes effort, but several agents pursuing the same open question at once ($\sum_{j\neq i}P_j(q,t)$) can on occasion help. Parallel attempts can cross-check and protect against single agent failure. The framework captures this by giving the two parts of $\Rred$ separate weights, so peer overlap can be rewarded ($\omega_p<0$) even while stock overlap is penalized.

Other extensions require new state variables. Adding a confidence or reliability term alongside $S_t$ may be useful for further verification, and if applied to machine intelligence, parallel compute may call for an explicit time-to-resolution variable, distinct from the private cost $C_i$.  

\section{Relation to prior work}\label{sec:prior}

The framework is inspired by growth macroeconomics. \citet{arrow1962} argued that inventors cannot capture all the rewards related to or produced by their discoveries. Here, that uncaptured social value appears as $\kappa_i<1$ inside an agent's value function. The shared knowledge stock is related to \citet{romer1990}, where knowledge is a non-rival input whose growth raises collective returns. 

\citet{weitzman1998} suggested that discovery depends on finding new combinations of existing ideas. The current model does not consider recombination. It counts only reusable, non-redundant additions to knowledge stock and allows stock to depreciate as newer work supersedes older work through the staleness rate $\delta$ in Equation~\ref{eq:stock} \citep[cf.][]{aghion1992}. 

The generativity term is directly inspired by \citet{murayama2022}, whose key insight is that learning can open new gaps faster than it closes old ones, so acquiring knowledge can make an agent aware of more unanswered questions. Murayama demonstrates this with a deliberately random model in which knowledge items form a network in which each is linked to a fixed fraction of the others, and the learner acquires them in random order. Under those assumptions, the number of open gaps rises while the learner knows less than half the items and only later declines. Differentially, in the model in the current manuscript, generativity $g_i$ is not a uniform property of interchangeable items but varies from question to question. Also, each agent selects a question according to its own (drifting) curiosity policy. 

Finally, the drift component of this framework converges with the effort recalibration framework of \citet{wiradhany2026}. That work was developed concurrently to address engagement with digital media and its impact on cognition. Both proposals concentrate on experience-dependent shifts of decision weights as an explanation for changes in human-technology interaction. These are modern, mechanistic forms of a concern voiced in Plato's \emph{Phaedrus}, in which Socrates warns that writing would reshape human memory and understanding. Importantly, the two papers differ in scope.  \citet{wiradhany2026} concerns itself with a choice between media engagement and effortful tasks, such that the weight of effort changes with repeated low-effort experience (e.g., scrolling social media). In the present framework, drift acts on all of the decision weights that govern a curiosity policy, of which effort is only one. It is embedded, moreover, in a multi-agent system with a shared knowledge stock, so that individual-level recalibration has population-level consequences for inquiry volume, diversity, and frontier share. Importantly, here, unlike a single-weight effort account, the same operator can move weights in either direction depending on the ecology. Therefore, the drift here is not always a downward slide toward low-effort or a maladaptive behavioral regime, it be can be adaptive or maladaptive in the nature depending on context (Section~\ref{sec:drift}). A companion paper (under review at \emph{Trends in Neurosciences}) extends the framework to neurobiological mechanisms and their broader sociological implications.

\section{Concluding remarks and limitations}\label{sec:open}

An important empirical question is whether curiosity weights drift in a manner that is captured by the framework. One could test this in an experiment which manipulates a person's recent history of inquiry and then measures which questions they choose over the following days. The drift term $\Delta\theta_i$ should eventually be tied to reinforcement learning. This requires identifying which signals update the weights, and over what timescales. 

The shared knowledge stock is a simplified scalar index $S_t$, paired with a set $\Sset_t$ of deposited results. But quantities such as the redundancy $d(q,\Sset_t)$ and the reusable value $U(q,\Sset_t)$ ultimately depend on the specific content of what is known and on how individual pieces of knowledge relate to one another. Therefore, the framework needs to be expanded (e.g., to represent knowledge in a a graph form).

For all multi-agent systems, the division of labor and the form of messages passed between agents are key. Peer coupling may coordinate a group, but it may also homogenize it. These issues need empirical study: when does social drift help, when does it instead collapse diversity, and how much information should be shared, and in what form, under different regimes? 

Cheap resolution to questions may not always be inherently harmful. If lower cost frees time to pursue the hard questions, frontier share can stay constant or even rise as volume rises. However, if $\tau_i$ falls, the same freed effort can be redirected toward shallow inquiry. External rewards may partly oppose the effects of drift: if abundant answers depreciate while rare long-horizon questions become more valuable, agents that internalize social value through $\kappa_i$ or $\eta_i$ may maintain a higher weight over long-horizon returns. Also, reusable by-products of cheap inquiry can lower the cost of harder questions \citep{arrow1962} and add material to $S_t$ for later recombination \citep{weitzman1998}. In that case, the experience state can record speed and low cost alongside the new gaps and reusable tools the inquiry produced. Then $\tau_i$ need not fall. Whether a policy $\theta_i$ is adaptive depends on its fit to the ecology $M_t$, not only on a policy by itself. For example, high vigilance, low tolerance for open questions, or reassurance-seeking can be useful in some settings, but not in others. 

If curiosity weights do in fact drift, it becomes important to distinguish the \emph{empirical} drift operator, meaning how the preferences of individuals and populations actually move as a function of behavior and ecology, from the \emph{desirable} drift operator, meaning how they would need to move to achieve a given objective, such as expanding useful knowledge. To capture such macro-level effects, we need to study how individuals and populations are curious and how their preferences drift, and the structure of the surrounding ecology itself, which describes which questions and answers are available, what they cost, and how answers propagate through a population.

\section*{Declarations}
\begin{itemize}
\item \textbf{Funding:} NIH/NIMH R01MH128344, R01MH110594, R01MH116937, and Conte Center MH10643.
\item \textbf{Conflict of interest:} The author declares no competing interests.
\item \textbf{Materials availability:} Not applicable.
\item \textbf{AI usage:} AI tools were used in preparation of the manuscript but the author takes full responsibility for the statements made herein.
\item The author would like to acknowledge Drs.\ Gaia Tavoni, Ethan Bromberg-Martin, Domenico Giannone, Bruno Averbeck, Michael Frank, Catherine Hartley, and Binxu Wang for many helpful comments on this article.
\end{itemize}

\bibliographystyle{apalike}
\bibliography{references}

\end{document}